\documentclass[a4paper,fleqn]{cas-dc}

\usepackage[numbers]{natbib}
\usepackage{ragged2e}
\usepackage{float}
\usepackage{indentfirst}
\usepackage[center]{caption}
\def\tsc#1{\csdef{#1}{\textsc{\lowercase{#1}}\xspace}}
\tsc{WGM}
\tsc{QE}
\tsc{EP}
\tsc{PMS}
\tsc{BEC}
\tsc{DE}

\begin{document}
\let\WriteBookmarks\relax
\def\floatpagepagefraction{1}
\def\textpagefraction{.001}
\shorttitle{Plastic Waste Segregation}
\shortauthors{Shivaank Agarwal et~al.}

\title [mode = title]{Application of Computer Vision Techniques for Segregation of Plastic Waste based on Resin Identification Code}                      

\author[1]{Shivaank Agarwal}

\ead{shivaank.agarwal@gmail.com}

\address[1]{Department of Computer Science, BITS Pilani,Hyderabad, India.}

\author[2]{Ravindra Gudi}

\ead{ravigudi@iitb.ac.in}

\address[2]{Department of Chemical Engineering, IIT Bombay, Mumbai, India}

\author[1]{Paresh Saxena}

\ead{psaxena@hyderabad.bits-pilani.ac.in}

\begin{abstract}
This paper presents methods to identify the plastic waste based on its resin identification code to provide an efficient recycling of post-consumer plastic waste. We propose the design, training and testing of different machine learning techniques to (i) identify a plastic waste that belongs to the known categories of plastic waste when the system is trained and (ii) identify a new plastic waste that do not belong the any known categories of plastic waste while the system is trained. For the first case, we propose the use of one-shot learning techniques using Siamese and Triplet loss networks. Our proposed approach does not require any augmentation to increase the size of the database and achieved a high accuracy of 99.74\%. For the second case, we propose the use of supervised and unsupervised dimensionality reduction techniques and achieved an accuracy of 95\% to correctly identify a new plastic waste. 
\end{abstract}


\maketitle

\section{Introduction}
\begin{figure}[ht]
\centering
\centerline{\includegraphics[width=85mm]{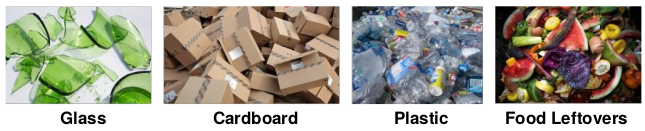}}
\caption{Different types of waste.}
\label{types of waste}
\end{figure}

\begin{figure}[ht]
\centering
\centerline{\includegraphics[width=85mm]{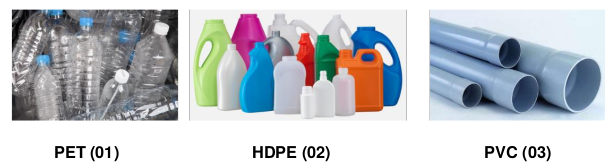}}
\caption{Different types of plastic waste.}
\label{types of plastic waste}
\end{figure}

As humans, we can visually identify and segregate various types of waste such as leftover food, metal cans, plastic bottles, cardboard boxes, and broken glass items as shown in Fig. \ref{types of waste}. All these categories of wastes are visually very different from each other and are recycled or disposed of based on their material. In the case of plastic waste, we usually segregate it from other waste, but we often do not realise that all plastics are not recycled the same way, and it is not visually possible for us to distinguish types of plastic as shown in Fig. \ref{types of plastic waste}. 

\vspace{-3mm}

\vspace{-3mm}

\begin{figure}[ht]
\centering
\centerline{\includegraphics[width=85mm]{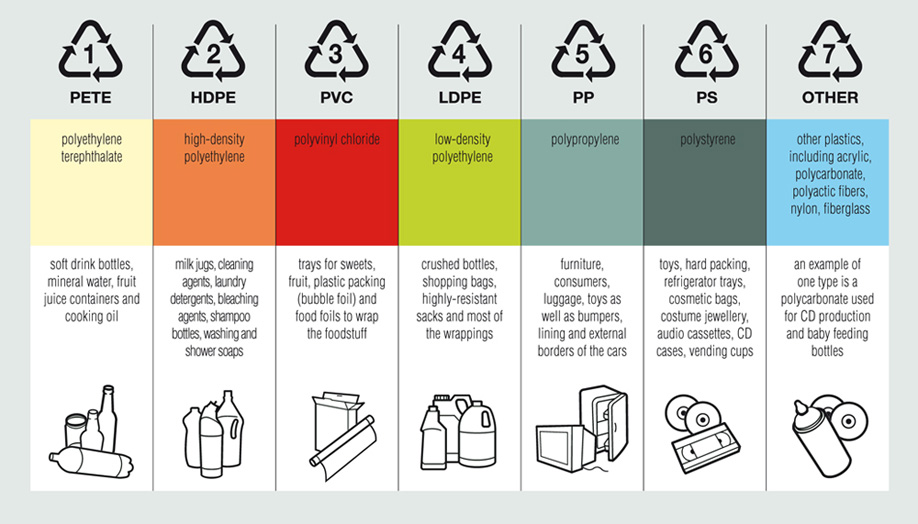}}
\caption{The 7 categories plastic resin codes.\cite{resin codes fig}}
\label{resin codes}
\end{figure}

The US society of the plastics industry introduced the Resin Identification Code (RIC) system in 1988, when the organization was called Society of the Plastics Industry, Inc. (SPI). The SPI states that the purpose of the plastic resin code is to provide a consistent system to facilitate the recycling of post-consumer plastic waste \cite{plastic}. The resin code is an indicator of the plastic resin out of which the product is made. There are a total of seven resin identification codes for plastic, as shown in Fig. \ref{resin codes}. Given an image containing a plastic waste, we aim to identify its resin code using image classification.

\vspace{-3mm}

\begin{figure}[t]
\centering
\centerline{\includegraphics[width=85mm]{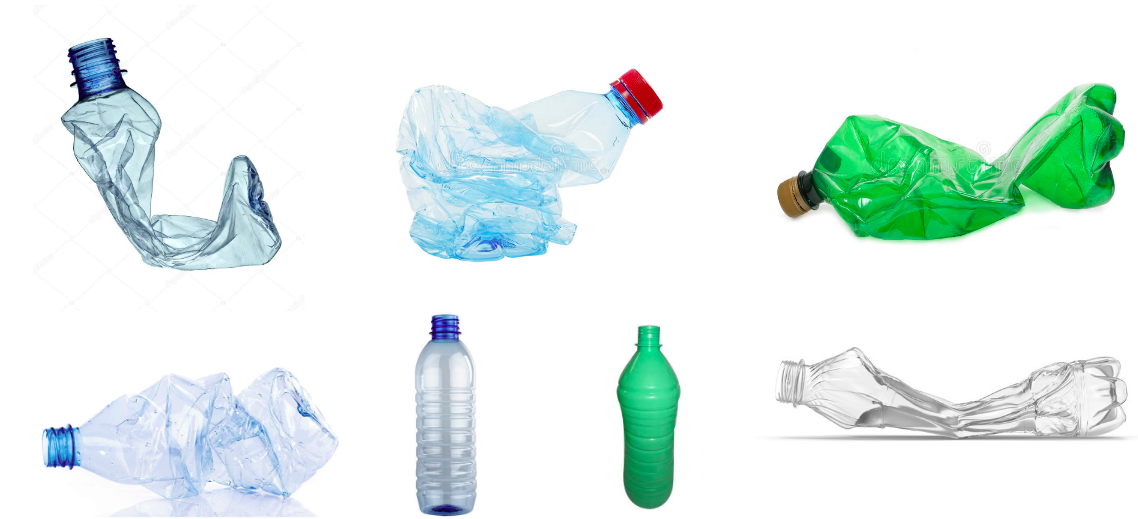}}
\caption{Plastic Bottles have a variety of colours, shape and textures}
\vspace{-3mm}
\label{bottles}
\end{figure}

Image classification is a supervised learning problem \linebreak where we classify an image into one of the predefined categories. It is different from an object detection problem since each image consists of a single object, and hence we do not need to localize the object. Earlier image classification algorithms relied on raw pixel data as the input. However, this approach does not give good results due to the difference in poses, angles, background, and lighting in different images for the same object category, as shown in Fig. \ref{bottles}. To identify objects more accurately, hand-engineered features such as edges, contours, colour histograms are extracted from images and used for the classification task. The problem with this approach is that the features have to be identified manually for each object category and it is challenging to identify and tweak features according to the problem statement. For example, in the case of classification of plastic bottles, there are several constraints: (i) to decide the colour to be taken into consideration, (ii) to know the shapes that represent bottles, (iii) variations of angles at which the bottles are photographed, etc.

 A breakthrough in solving the problem of image classification came with the use of convolutional neural networks (CNN) on image pixels \cite{lenet},\cite{vgg16}. Given input images and the corresponding labels, CNNs automatically learn the features require to classify an image. With the use of CNNs, deep learning has achieved better results in tasks such as image classification and object detection \cite{yolo,ssd}. However, these algorithms require a large amount of data for training, often in the range of hundreds of thousands. With a smaller amount of data, deep neural networks tend to overfit on the training data and hence they are not able to generalize on the entire dataset. In such a case, it is better to learn discriminative features to distinguish between classes. 
 
\begin{figure}[h]
\centering
\centerline{\includegraphics[width=70mm]{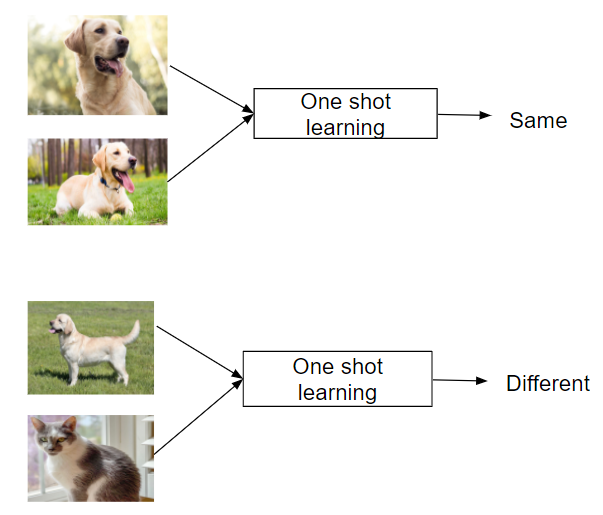}}
\caption{Demonstration of One Shot Learning technique}
\vspace{-3mm}
\label{one shot learning}
\end{figure}

 In this paper, we propose the use of one-shot learning techniques \cite{one shot learning 1,one shot learning 2} to train the network over smaller amounts of data for plastic waste classification. One-shot learning is a classification task where only a small number of examples are given for each class. The methods used to solve one-shot classification tasks directly learn similarities between the same class objects and the differences between objects of different classes. During training, these methods select two or three instances of data points during a single forward pass of the network as shown in Fig. \ref{one shot learning}. Hence from a data set of size $n$, a total of $n^2$ training instances are generated, which is useful if the size of the data set is small.

Specifically, our proposed approach investigates the use of one-shot learning techniques with Siamese \cite{Koch} and Triplet loss network \cite{facenet} for plastic waste classification. These methods have shown excellent results on tasks such as hand-written character classification and face detection \cite{Koch,facenet}. In such cases, the number of instances per class is small and the number of classes is large. Further, we have also proposed techniques to classify plastic waste that do not belong to the known plastic waste categories using supervised and unsupervised dimensionality reduction techniques. We demonstrate our results with the WaDaBa database \cite{WaDaBa1} that consists of multiple images of plastic waste divided into several categories. In this paper, our main contributions are as follows.

\begin{itemize}
    \item First, we propose the use of one-shot learning techniques with Siamese and Triplet loss network for plastic waste classification. We extract features for each image in such a way that the distance between the resultant image embedding provide the indication of their similarity. We then classify an image based on its distance from the embedding of images belonging to a particular class. Our proposed approach achieves an accuracy of 99.74\% on the database. Additionally, our proposed approach does not require any augmentation to increase the size of the database and at the same time achieves a high accuracy.
    
    \item Second, we have proposed the techniques of supervised and unsupervised dimensionality reduction to visualize the data and form clusters of classes for classification of plastic waste of unknown category. Based on a data point's position relative to other points in a cluster, we classify the point as a new plastic or as belonging to an already known class. Our proposed approach achieves 95\% in correctly identifying a new plastic as an outlier.
    
\end{itemize}

\vspace{2mm}
This paper extends the work in \cite{shivaank}, where we have presented a brief overview of the plastic waste classification when all the categories are known. In this paper, we have extended the work in two directions. First, we describe in detail the complete procedure to classify the plastic waste data into a known category and second, we have also proposed procedures to identify the new plastic waste that does not belong to any of the previously seen categories.
 
The rest of the paper is described as follows. In Section 2, we describe the WaDaBa database and review the work previously done using this database. In Section 3, we describe the preliminaries on convolutional neural networks. Section 4 presents our proposed approach and description of our proposed architectures with Siamese and Triplet loss networks. It also explains the procedure to classify a new plastic waste using dimensionality reduction techniques. In section 5, we present the training details of the proposed approaches. Section 6 presents the methods used to evaluate our algorithm and the results obtained subsequently. Finally, we conclude this paper in Section 7.

\section{Related Work}

There are various datasets \cite{trashnet,taco} on waste classification which can be used to segregate plastic from other types of waste such as cardboard, glass, and paper. WaDaBa \cite{WaDaBa1} is the only available dataset that classifies plastic waste based on its resin code. It consists of 4000 images belonging to the five categories of plastic resin codes - PET (01-polyethylene terephthalate), PE-HD (02-high-density polyethylene), PP (05-polypropylene), PS (06-polystyrene), and other (07). Table \ref{image categ} shows the number of images per category in the database. Each image consists of a single object belonging to one of the above classes. The objects used in the database are acquired from households over a period of four months to incorporate a variety of municipal solid waste.

\begin{center}
\begin{table}[pos=h]
\caption{Images per category in WaDaBa}
\begin{tabular}{@{} l c @{}}
\hline
\textbf{Category} & \textbf{No. of Images} \\ \hline
01 PET & 2200 \\ 
02 PE-HD & 600 \\ 
05 PP & 640 \\ 
06 PS & 520 \\ 
07 Other & 40 \\
\hline
\end{tabular}
\label{image categ}
\end{table}
\end{center}

\begin{center}
\begin{figure}[h]
\centerline{\includegraphics[width=70mm]{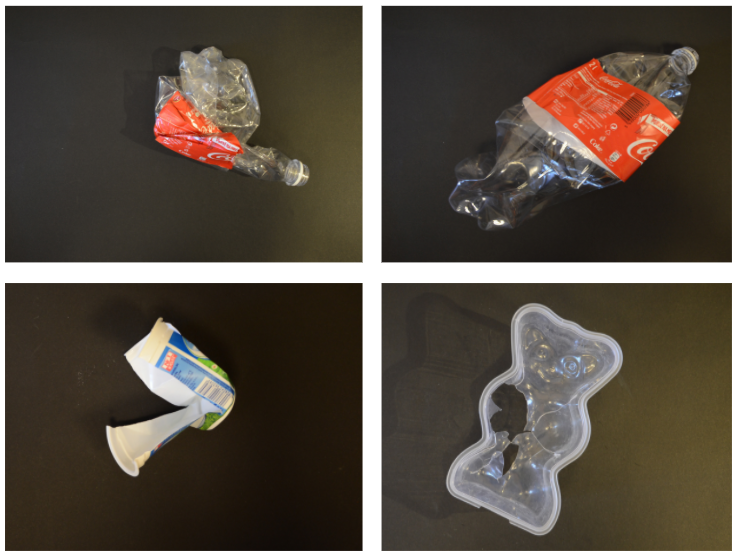}}
\caption{Sample images from the WaDaBa database. Top-left: crushed plastic bottle. Top-right: Same bottle subjected to another deformation. Bottom-left: Torn plastic container. Bottom-right: Plastic container with missing parts}
\label{wadaba images}
\end{figure}
\end{center}

The objects in the image are subjected to different lighting conditions, various degrees of damages, and are rotated at different angles to simulate natural conditions. Few images in the database are also photographed at odd angles and have varying degrees of deformations as shown in \mbox{Fig. \ref{wadaba images}.}

There have been two previous works on this database. In \cite{WaDaBa1} the authors use histogram analysis to classify PET\linebreak( Category 01 in Fig. \ref{resin codes}) from non PET waste. They achieved an average accuracy of 75.68\% on the dataset. While histogram analysis is computationally efficient compared to deep neural networks, the accuracy of classification is low for industrial standards. Moreover, using histogram analysis it is only possible to classify a plastic as PET or non- PET and not into other categories such as PP, PS and PE-HD. In \cite{WaDaBa2} the authors have used deep convolutional neural networks to classify waste for four out of the five available categories in the WaDaBa database. They have used image augmentation (mainly rotation) to increase the size of the database from 4000 to 140,000 images. They achieved an accuracy of 99.92\% on the four categories. While a high accuracy is achieved, the above method requires substantial image augmentation (more than 30 times the original data). The data is first augmented and then split into train and test sets. This procedure does not add to any unseen data in the test class since CNN models are powerful enough to learn rotational invariant filters, given a sufficient amount of augmented training data \cite{rotational invariance}.  Hence the above method does not work well on new instances of data that do not have a rotational variant in the training set. In this paper, we propose a  model that does not require any augmentation to increase the size of the database but also achieves a high accuracy. In addition, we have also proposed the solution to classify a plastic material that may not belong to any of the known categories.

\section{Preliminaries}
\begin{figure}[h]
\centering
\centerline{\includegraphics[width=70mm]{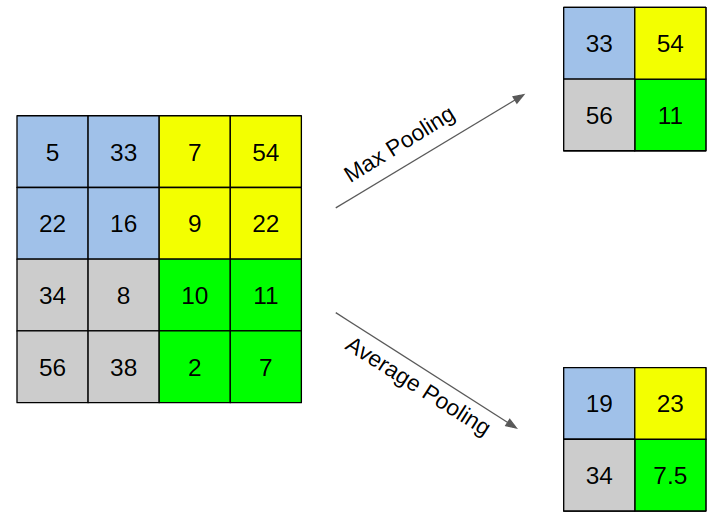}}
\caption{Example of a pooling operation}
\vspace{-1mm}
\label{pooling}
\end{figure}

CNNs consist of a stack of modules that perform three operations on the input image namely convolution, rectified Linear units (ReLU), and pooling. These operations are explained as follows.
\begin{itemize}
    \item {Convolution}: During the convolution operation, filters are applied to the input image (or previous layer) to produce a new layer called the output layer, which may have a different height, width, or depth than the input layer. The depth of the filter is equal to the depth of the input layer. The output layer is computed by sliding the filters across the input and performing element-wise multiplication. After multiplication, all the resultant layers are added to produce a single feature map. This is done for all the filters in that particular layer. Hence the number of output feature maps is equal to the number of filters. Fig. \ref{convolution} shows an example of the convolution operation with input $x$ (7x7x3), two filters $w_{0}$ (3x3x3) and $w_{1}$ (3x3x3), and output $o$ (5x5x2). 

\begin{figure}[t]
\centering
\centerline{\includegraphics[width=85mm]{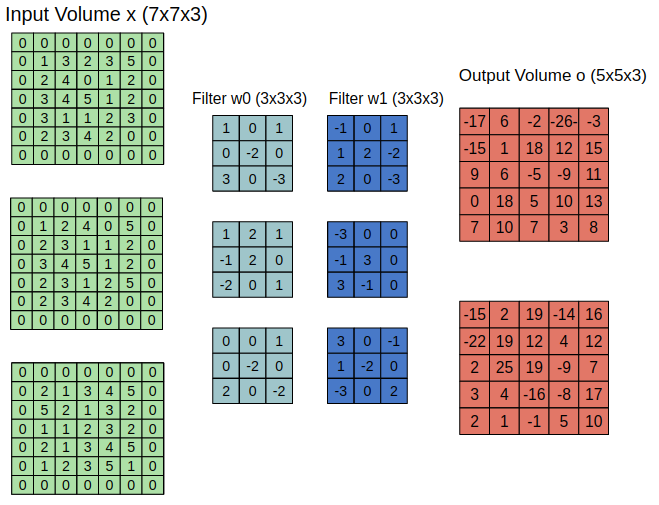}}
\caption{Example of a convolution operation.}
\vspace{-3mm}
\label{convolution}
\end{figure}
\vspace{2mm}

\item{ReLU}: The ReLU function is applied to introduce non-linearity after the convolution operation. The ReLU function, $F(x) = max(0,x)$ , returns $x$ for all values of $x > 0$, and returns $0$ for all values of $x \leq 0$.

\item{Pooling}: The pooling is followed after ReLU. In pooling, the image is downsampled by preserving only the maximum or the average of values in a neighbourhood. This is done to reduce computation time.
Fig. \ref{pooling} shows a pooling operation where a 4x4 grid is downsampled to a 2x2 grid.

\end{itemize}

At the end of all the convolution layers, we have fully connected (FC) layers in which every neuron in the input layer is connected to every neuron in the output layer. The FC layers are used to classify the images based on the features extracted by the convolutional layers.

\section{Proposed Approach}
In this section, we describe the proposed approach to classify plastic waste. First, we describe the methods to classify plastic waste into pre-defined categories in Section \ref{sec: classification}, and then we describe the methods to identify plastic waste that does not belong to any of the known categories in Section \ref{sec: identification}.

\subsection{Classification of known categories}\label{sec: classification}

To achieve the first objective of classifying plastic waste of known categories based on its resin codes, we propose to use the Siamese and Triplet Loss convolutional neural networks.

\subsubsection{Siamese Network}

\begin{figure}[h]
\centerline{\includegraphics[width=85mm]{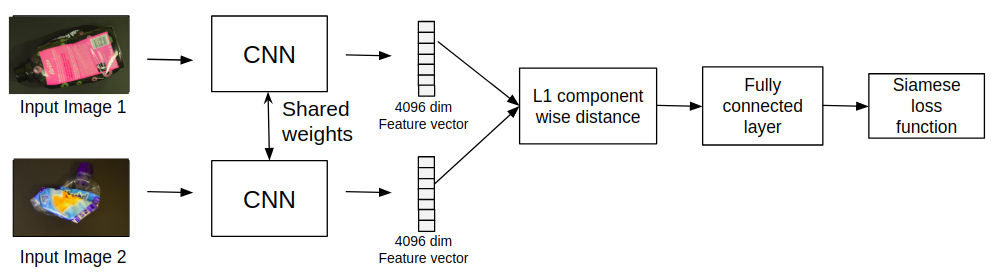}}
\caption{Proposed architecture to identify plastic waste using Siamese network structure.}
\label{Siamese}
\end{figure}

The Siamese network aims at learning image representations such that the distance between output vectors of images belonging to the same class is less than that of images belonging to different classes. The proposed architecture to identify plastic waste with the use of the Siamese network is shown in Fig \ref{Siamese}.

\begin{center}
\begin{figure*}[t,width=\textwidth]
\centerline{\includegraphics[width=140mm,height=30mm]{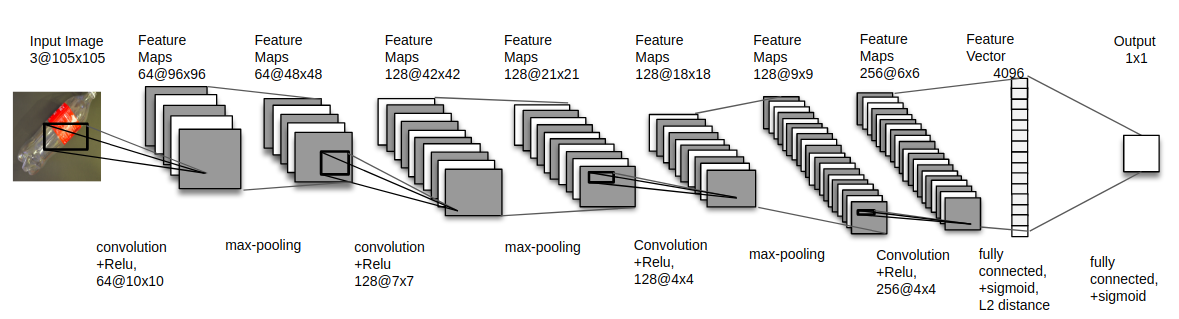}}
\caption{Siamese Convolutional Neural Network Structure. Siamese twin joins immediately after the 4096 unit fully-connected layer where the $L_{1}$ component-wise distance between vectors is computed.}
\label{cnn_Siamese}
\end{figure*}
\end{center}

It consists of 2 identical CNNs which have shared weights. Further, Fig \ref{cnn_Siamese} represents the overall system model of CNN which consists of 4 convolutional layers and two fully connected layers. The first three convolutional layers are each followed by a max-pool layer. The 2 networks calculate the respective embedding of size 4096 and merge together by calculating the $L_{1}$ component wise distance between the two embedding which results in a single vector of size 4096. This vector is then transformed into a single output of shape $1\times1$ by a fully connected layer. This output and the label are then fed into the cross entropy loss function as shown by equation \ref{eq_Siamese}. 

\begin{equation}
\begin{split}
 \\L_S(x^1,x^2) = y(x^1,x^2)logp(x^1,x^2) + \\
     (1-y(x^1,x^2))(1-logp(x^1,x^2)) \label{eq_Siamese}
\end{split}
\end{equation}

where $x^1$ and $x^2$ are the 2 images fed into the Siamese network and $p(x^1,x^2)$ is the output.
The label $y(x^1,x^2)$ = 0 if $x^1$ and $x^2$ belong to the same class else $y(x^1,x^2)$ = 1. The loss calculated using equation 1 is propagated backwards in both the identical networks.

\subsubsection{Triplet loss Network}

\begin{figure}[h]
\centerline{\includegraphics[width=70mm]{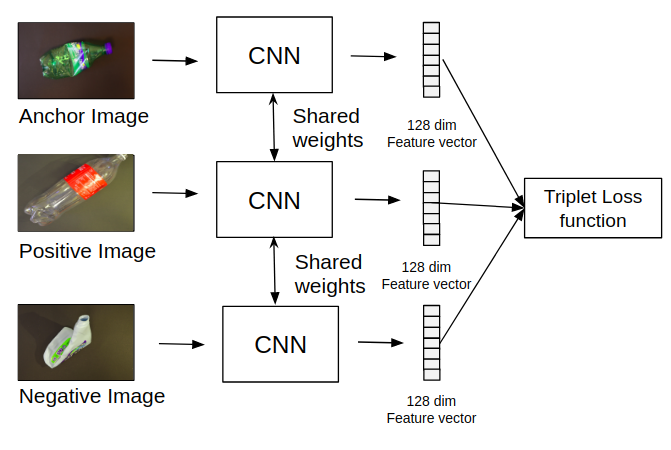}}
\caption{Triplet Network Structure}
\label{Triplet}
\end{figure}

\begin{center}
\begin{figure*}[t,width=\textwidth]
\centerline{\includegraphics[width=140mm,height=30mm]{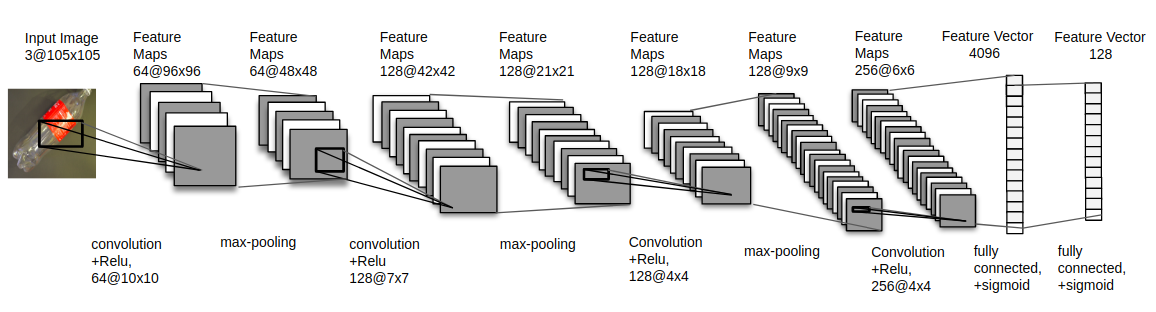}}
\caption{Triplet Loss Convolutional Neural Network Structure. The other two parallel CNNs are not depicted, but they join immediately after the 128 unit fully-connected layer where all the 3 embedding are fed into the Triplet loss function.}
\label{cnn_Triplet}
\end{figure*}
\end{center}

The Triplet loss network also aims at learning image embeddings such that the distance between image embeddings of plastics belonging to the same category is less compared to that of the plastics belonging to different categories. The proposed Triplet Loss network structure is shown in \mbox{Fig. \ref{Triplet}}. The main difference between Triplet and Siamese network is that the Triplet loss network consists of 3 identical convolutional neural networks which take 3 images as inputs during a single forward pass of the network. Two images are known as the anchor image and positive image. They belong to the same class while the third image, known as the negative image, belongs to a different class. The loss function shown in \mbox{equation \ref{eq_Triplet}} is constructed in such a way that the image embeddings of a particular class is enforced to be closer to all image embeddings of the same class than it is to the embeddings of another class. A constant margin of 0.4 is used to increase the distance between embeddings of images belonging to different classes.

\begin{equation}
\begin{split}
\\ L_T(x^a,x^p,x^n) = ||f(x^a) - f(x^p)||^2_2  \\ -||f(x^a) - f(x^n)||^2_2 + 0.4 \label{eq_Triplet}
\end{split}
\end{equation}
where $f(x^a)$ is an anchor image embedding, $f(x^p)$ is a positive image embedding, $f(x^n)$ is a negative image embedding and $||$  $||^2_2$ denotes square of the euclidean norm. The CNN architecture for the Triplet loss network is \linebreak shown is Fig. \ref{cnn_Triplet}. The image embedding in this case is of 128 dimensions (number of neurons in the last fully connected layer). For training, 3 images are taken as the input in which 2 belong to the same class. Just like the Siamese, the images can be paired in any order since the architecture and weights of the 3 CNNs are identical. The loss calculated subsequently is propagated backwards in all the 3 identical networks. 

\subsection{Identification of a new plastic waste} \label{sec: identification}
In the previous subsection, we proposed methods to classify plastic waste from the known categories. However, there is a possibility that we encounter a plastic made up of an unknown resin. In such a case, the plastic does not fall into any of the known categories. Hence in this subsection, we propose the method to identify a plastic waste of unknown category.

Note that since we do not have any data about the plastics with unknown categories, we assume that we do not encounter any of these during the training phase. Hence traditional neural networks cannot be used since they classify data into already known classes. Further, zero-shot learning can be used to classify instances of data that have not been seen during the training phase \cite{zero shot}, however, the prerequisite of this approach is that classes need to have a predefined structure. For example, in the case of animals, these descriptions could include their stripes, number of legs, and colour. In our case, we have no information about the types of plastics we might encounter during the test phase. Therefore, zero-shot learning is not a feasible solution to solve the problem.

In this paper, we formulate the problem of identifying a new plastic waste by treating a new category of plastic as an outlier detection problem. We use the architecture described in Triplet loss network as our base model. We train it on 4 of the 5 categories and treat the 5th category as a new plastic. We consider only one category as a new plastic category since the larger the number of known categories, the greater is the probability that an unknown plastic may be similar to one of the known categories. Hence to maximize the variation in known plastics, we chose only one class as the class of new plastics. After training, we extract the features from the fully connected layers for all the known and unknown categories. We then perform classification based on supervised and unsupervised dimensionality reduction as described in the next subsections.

\subsubsection{Classification based on unsupervised dimensionality Reduction (PCA)}

Principal Component Analysis (PCA) is one of the most widely used techniques for dimensionality reduction \cite{pca}, \cite{pca2}. PCA projects data into multiple number of dimensions. However, the number of dimensions cannot exceed the minimum of the number of features and the number of data points. The direction vectors on which the data is projected are the eigenvectors of the data’s covariance matrix. The direction vectors are chosen this way because they are orthogonal to each other, and they retain the maximum possible variance of the data in that dimension. PCA is an unsupervised technique since it does not take into account the labels of the data points. In our approach, we project the data belonging to the training class on various dimensions using PCA and then use the resultant direction vectors to project the test data.

\subsubsection{Classification based on supervised dimensionality reduction (LDA)}
Linear discriminant analysis (LDA) is a predictive algorithm for multi-class classification \cite{lda}.It is also used for dimensionality reduction of data. LDA is used to find a linear combination of features to discriminate between classes. It computes direction vectors that maximize the distance between classes while simultaneously reducing the within-class variance. Since LDA uses class labels to compute the projections of higher dimension points to lower dimensions, it is a supervised algorithm. In our approach, we first use the training data to compute the direction vectors using LDA and then we project the test data in these direction vectors to classify them as an outlier.

\section{Training and Preprocessing}
In this section, we explain the training procedures used to classify known and unknown categories.

\subsection{Classification of known categories}
Before training, the images are resized to a height and width of $105\times105$ with three channels (RGB). The dataset is normalized across all three channels by calculating the mean and variance of all the plastic waste images in the dataset. For both the Siamese and Triplet loss networks, data is split in an 80:10:10 ratio for the training, validation, and test set respectively. Therefore, the training, validation, and test set consist of 3200, 400 and 400 images respectively. The data split for each class is proportional to the size of the class.

\subsubsection{Siamese Network}
\begin{figure}[h]
\centerline{\includegraphics[width=80mm]{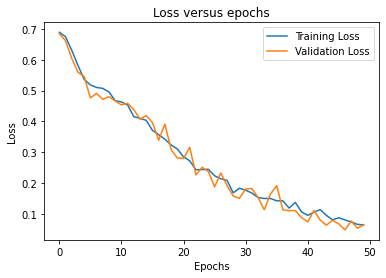}}
\caption{Training and validation loss for the Siamese network}
\label{Siamese loss}
\end{figure}

The image pairs are created by choosing a random image from the test set and then choosing another random image either from the same class or from another class in the test set itself where 50\% of the image pairs consist of images from the same class. The Siamese network is trained for 50 epochs. In each epoch, 5000 image pairs are generated from the test data. A batch size of 50 is used for training. Stochastic gradient optimizer is used with a learning rate of 0.001 and momentum of 0.9. Fig. \ref{Siamese loss} shows training and validation loss for the Siamese network with respect to the number of epochs where the loss converges with the increase in the number of epochs.

\subsubsection{Triplet loss Network}
\begin{figure}[h]
\centerline{\includegraphics[width=80mm]{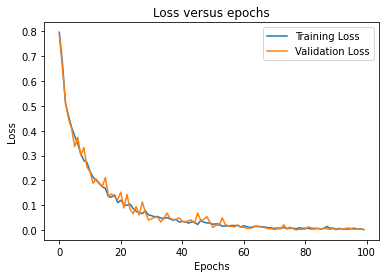}}
\caption{Training and validation loss for the Triplet loss network}
\label{Triplet loss}
\end{figure}

The image Triplets are created by choosing a random image from the test set (anchor image), another random image from the same class (positive image) and a random image from a different class (negative image). All three images are chosen from the test set itself. The Triplet loss network is trained for 100 epochs. In each epoch, 5000 image Triplets are generated from the test data. A batch size of 50 is used for training. Stochastic gradient optimizer is used with a learning rate of 0.001 and momentum of 0.9. \mbox{Fig. \ref{Triplet loss}}  shows training and validation loss for the Triplet loss network with respect to the number of epochs. Our results show that the loss also converges with the increase in the number of epochs in this case as well.

\subsection{Identification of new plastic waste} 

For pre processing the data, the entire dataset is resized and normalized as done in the classification of known categories. The category of other plastic (Category - 07) is treated as new plastic, and hence it is not included in the training set. we select category 07 as the class of new plastics since it has the least number of images, and in real life, we would rarely encounter unknown classes compared to the known classes. The rest of the four categories is divided into a ratio of 80:10:10 for the train, validation, and test sets. The training procedure is the same as that of the Triplet loss network. After training, the network is used to extract features of the entire dataset from the last fully connected layer before the output from the three identical networks are fed into the Triplet loss function. Hence each image is represented by a vector of size 128. For the task of outlier detection, the features extracted from the training data are used by PCA and LDA to learn the direction vectors for the projection of test data. The test data is then projected on these vectors and classified as an outlier based on its relative distance from neighbouring points. The exact algorithm is discussed in the verification and results section.

\section{Verification and Results}
In this section, we demonstrate the benefits of using our proposed methods for the classification of plastic waste for both known and unknown categories.
\subsection{Classification of known categories}
To demonstrate the classification of plastic into one of the five known categories, we have used the following two verification methods:

\subsubsection{N-shot K-way accuracy} In this method, $N$ stands for the number of labels or image classes (for example, $N=5$ in our case), and $K$ stands for the number of images per label. Given $K$ images of each of the $N$ labels, we need to classify the test image into one of the $N$ categories. We do this using only one image per class (i.e., $K=1$). For each of the 400 test images, we select 5 random images, one from each class. We form 5 pairs, each consisting of the selected test image and the image chosen at random. For the Siamese network, all the 5 image pairs are sent as input into the network. The test image is assigned to the class for which the image pair has the least output. In the Triplet loss network, the 128 dim features are extracted from the test image and all the other 5 images. The Euclidean distance between the 5 image pairs is calculated, and the test image is assigned to the class with whose image embedding it have the least distance. In both the Triplet loss and Siamese networks, the count of correctly classified images is incremented if the assigned class matches with the original label of the test image. Table \ref{one shot accuracy} shows the 1-shot 5-way accuracy obtained on the 400 test images. Our results show that accuracy of 99.50\% and 99.0\% is achieved by using Siamese and Triplet loss networks, respectively.

\begin{table}[pos=h]
\caption{1-Shot 5-way  accuracy for the classification of plastic waste of known categories using Siamese and Triplet loss networks.}
\begin{center}
\begin{tabular}{@{} c|c @{}}
\hline
\textbf{Method} & \textbf{Accuracy} \\ \hline
Siamese & 99.50\\
Triplet Loss & 99.0\\
\hline
\end{tabular}
\end{center}
\label{one shot accuracy}
\end{table}

\subsubsection{K-Nearest Neighbour} In this method, we first extract embedding of all the images in the dataset using the Siamese and Triplet loss networks. In the case of Siamese, the embedding size is 4096, while in the case of Triplet Loss, the size is 128. We then compute the Euclidean distance between the embedding of the test image and all the other images. Next, we apply the $K$-nearest neighbour algorithm by assigning the test image to the category having the majority number of images in the top $K$ closest matches. The experiment is repeated for various values of $K$, and the results are shown in Table \ref{Siamese knn} and Table \ref{Triplet knn}. Our results show that a high average accuracy of 99.72\% and 99.79\% is achieved by using Siamese and Triplet loss network, respectively.  

\begin{table}[pos=h]
\addtolength{\parskip}{-1mm}
\caption{KNN Accuracy for Siamese Network}
\begin{center}
\begin{tabular}{@{} c|c|c|c|c|c|c @{}}
\hline
\textbf{K} &\textbf{PET} & \textbf{PE-HD}
&\textbf{PP} & \textbf{PS} & \textbf{Other} & \textbf{Average} \\ \hline
3 & 99.59 & 99.66 & 99.35 & 100 & 100 & \textbf{99.72}\\
5 & 99.4 & 99.83 & 99.00 & 100 & 100 & 99.64\\
7 & 99.59 & 99.66 & 99.20 & 100 & 100 & 99.69\\
\hline
\end{tabular}
\end{center}
\label{Siamese knn}
\end{table}

\begin{table}[pos=h]
\addtolength{\parskip}{-1mm}
\caption{KNN Accuracy for Triplet Loss Network}
\begin{center}
\begin{tabular}{@{} c|c|c|c|c|c|c @{}}
\hline
\textbf{K} &\textbf{PET} & \textbf{PE-HD}
&\textbf{PP} & \textbf{PS} & \textbf{Other} & \textbf{Average} \\ \hline
3 & 99.68 & 100 & 99.30 & 100 & 100 & \textbf{99.79}\\
5 & 99.50 & 100 & 98.5 & 98.5 & 100 & 99.56\\
7 & 99.40 & 100 & 98.40 & 99.8 & 100 & 99.52\\
\hline
\end{tabular}
\end{center}
\label{Triplet knn}
\end{table}

\subsection{Identification of new plastic waste} 
In both the methods (PCA, LDA), we first use the training data to calculate the direction vectors and then project the test data on the calculated vectors. The test data consists of the entire class of new plastic (Category 07) and 10\% of data from the rest 4 classes. Hence the test data consists of a total of 436 image embeddings. After projection, we calculate the number of data points within a certain distance (say X) of the projected embedding. If the number of such data points is less than a threshold (say Y), we classify the image as an outlier. X and Y are hyper-parameters which are chosen by trial and error using the validation set. We evaluate our results by calculating the number of true positives (TP), false positives (FP), true negatives (TN), and false negatives (FN). Each of these four measures represent the following:

\begin{itemize}
    \item TP - Number of new plastics predicted as outliers.
    
    \item FP - Number of known plastics predicted as outliers.
    
    \item TN - Number of known plastics predicted as interior points.
    
    \item FN - Number of new plastics predicted as interior points.
    
\end{itemize}

Further, we measure the two algorithms by using the following two criteria. First, the percentage of new plastic that is correctly identified as an outlier i.e. TP/TP+FN. This quantity should be as high as possible since we want to decrease the number of new plastics segregated into one of the known categories. Second, the percentage of known plastic that is wrongly classified as an outlier i.e. FP/FP+TN. This quantity should be minimal since we want to decrease the number of known plastics segregated as new plastic. For each of the two methods(PCA and LDA), we show the results obtained after projection to various dimensions. We also show the plot after projection to two dimensions for a visual representation of the projected data.

\subsubsection{Identification of a new plastic waste using PCA}
\begin{figure}[h]
\centerline{\includegraphics[width=60mm]{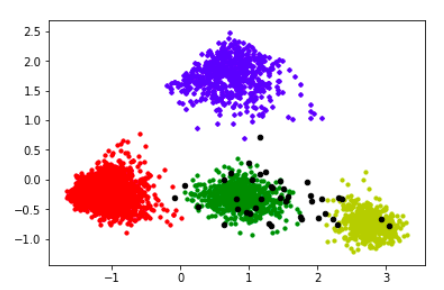}}
\caption{Principal Component Analysis in 2 dimensions for the identification of a new plastic waste. The black colour data points represent the new plastic waste.}
\label{pca}
\end{figure}

Table \ref{pca accuracy} shows the accuracy for multiple dimensions for PCA. For the illustration purposes, Fig. \ref{pca} shows the plot obtained after projection to two dimensions using PCA. Note that in Fig. \ref{pca}, the colours red, blue, green, and yellow represent the data points belonging to classes PET, PE-HD, PP, and PS, respectively. The colour black represents the data from the class of other plastics, which are treated as new plastic. The maximum accuracy is obtained when the data is projected to three dimensions where 87.5\% of new plastic is correctly identified as an outlier and only 12.12\% of known plastic is wrongly categorised as a new plastic. Increasing the number of dimensions beyond three does not help in improving the classification accuracy since they may act as noise.

\begin{table}[pos=h]
\addtolength{\parskip}{-1mm}
\caption{PCA Accuracy with different dimensions to identify a new plastic waste}
\begin{center}
\begin{tabular}{ |c|c|c|c|c| }
\hline
\textbf{Dimension} & \textbf{TP } &\textbf{FP} & \textbf{TN}
&\textbf{FN} \\ \hline
1 & 33 & 76 & 320 & 7 \\ \hline
2 & 34 & 60 & 330 & 6 \\ \hline
3 & \textbf{35} & \textbf{48} & \textbf{348} & \textbf{5} \\ \hline
4 & 35 & 69 & 327 & 5 \\ \hline
5 & 34 & 60 & 336 & 6 \\ 
\hline
\end{tabular}
\end{center}
\label{pca accuracy}
\end{table}

\subsubsection{Identification of a new plastic waste using LDA}
\begin{figure}[h]
\centerline{\includegraphics[width=60mm]{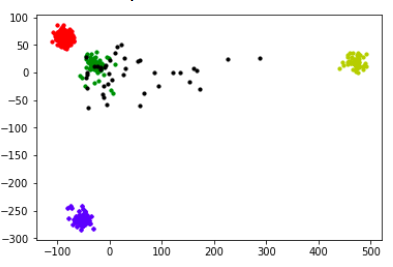}}
\caption{Linear Discriminant Analysis in 2 dimensions. The black colour data points represent the new plastic waste.}
\label{lda}
\end{figure}

Finally, Table \ref{lda accuracy} shows the accuracy achieved by LDA for multiple dimensions. Similarly, Fig. \ref{lda} shows the plot obtained after projection oto two dimensions using LDA where colors represent different categories as mentioned in the previous subsection. Since LDA generates $C-1$ discriminant planes where $C$ is the number of classes, we cannot project the data to more than $C-1$ dimensions. In our case the number of training classes is 4, and hence we show the results obtained for projection to 1,2 and 3 dimensions. The highest accuracy is achieved when the data is projected to three dimensions, where 95\% of new plastic is correctly identified as an outlier and only 8.6\% of known plastic is wrongly categorised as new pastic.

\begin{table}[pos=h]
\addtolength{\parskip}{-1mm}
\caption{LDA Accuracy with different dimensions to identify a new plastic waste}
\begin{center}
\begin{tabular}{ |c|c|c|c|c| }
\hline
\textbf{Dimension} & \textbf{TP } &\textbf{FP} & \textbf{TN}
&\textbf{FN} \\ \hline
1 & 32 & 72 & 324 & 8 \\ \hline
2 & 34 & 45 & 351 & 6 \\ \hline
3 & \textbf{38} & \textbf{34} & \textbf{362} & \textbf{2} \\ 
\hline
\end{tabular}
\end{center}
\label{lda accuracy}
\end{table}

\section{Conclusion and Future Work}
In this paper, we have proposed methods to solve two problems. First the identification of a plastic waste from the known categories of plastic waste when the system is trained. We have achieved a maximum accuracy of 99.79\% on the WaDaBa database across all the 5 plastic categories without any image augmentation. Second, we are also able to distinguish any new plastic waste with a accuracy of 95\% while wrongly classifying the known plastics only 8.6\% as a new plastic. Future work will focus on object localization followed by the classification techniques specifically in case of multiple waste objects within a single image.

\end{document}